%% file: acl2023.tex
\pdfoutput=1

\documentclass[11pt]{article}

\usepackage{ACL2023}

\usepackage{times}
\usepackage{latexsym}

\usepackage[T1]{fontenc}

\usepackage[utf8]{inputenc}

\usepackage{microtype}

\usepackage{inconsolata}

\usepackage{amsmath}
\usepackage{url}
\usepackage{xurl}

\usepackage{graphicx}
\usepackage{amsfonts}
\usepackage{CJK}
\usepackage{bm}
\usepackage{mathrsfs}
\usepackage{float}
\usepackage{xcolor,colortbl}

\usepackage{subfigure}
\usepackage{arydshln,booktabs,multirow}
\usepackage{bigstrut,bigdelim}
\usepackage{paralist}
\usepackage{bbm}

\usepackage{helvet}
\usepackage{courier}
\usepackage{makecell}
\usepackage{nicefrac}
\usepackage{epsfig}
\usepackage{diagbox}
\usepackage{framed}
\usepackage{empheq}
\usepackage{array}
\usepackage{enumitem}
\usepackage{mdwlist}
\usepackage{xspace,mfirstuc,tabulary}
\usepackage{tcolorbox}
\usepackage{algorithm}
\usepackage{amsmath}
\usepackage{amssymb}
\usepackage{mathtools}
\usepackage{amsthm}
\usepackage{placeins}
\usepackage{algpseudocode}

\DeclareMathOperator*{\softmax}{softmax}

\DeclareMathOperator*{\TransformerEncoder}{CE}
\DeclareMathOperator*{\CACE}{CACE}
\DeclareMathOperator*{\TransformerDecoder}{TD}
\DeclareMathOperator*{\CA}{Cross-Attention}

\newcommand{\ModelName}{DiffuSIA}

\newcommand{\defeq}{\coloneqq}

\newcommand{\bx}{\mathbf{x}}

\newcommand{\bmu}{{\boldsymbol{\mu}}}

\newcommand{\exper}[1]{\textsc{#1}}
\newcommand{\expmb}[1]{\mathbf{#1}}
\newcommand{\expobj}[1]{\mathcal{L}_\text{#1}}
\newcommand{\xx}{\expmb{x}}

\newcommand{\ww}{\expmb{w}}
\newcommand{\cc}{\expmb{c}}
\newcommand{\cx}[1]{\mathbf{x}_{#1}}
\newcommand{\ptheta}{p_\theta}

\newcommand{\qphi}{q_\phi}
\newcommand{\Lsimple}{\expobj{simple}}

\newcommand{\Emb}{\exper{Emb}}

\title{\textsc{DiffuSIA}: A Spiral Interaction Architecture for \\ Encoder-Decoder Text Diffusion}

\author{Chao-Hong Tan, Jia-Chen Gu, Zhen-Hua Ling \\
  National Engineering Research Center of Speech and Language Information Processing, \\
  University of Science and Technology of China, Hefei, China \\
{\tt chtan@mail.ustc.edu.cn}, {\tt \{gujc,zhling\}@ustc.edu.cn}
}

\begin{document}
\maketitle
\begin{abstract}
  Diffusion models have emerged as the new state-of-the-art family of deep generative models, and their promising potentials for text generation have recently attracted increasing attention.
  Existing studies mostly adopt a single encoder architecture with partially noising processes for conditional text generation, but its degree of flexibility for conditional modeling is limited.
  In fact, the encoder-decoder architecture is naturally more flexible for its detachable encoder and decoder modules, 
  which is extensible to multilingual and multimodal generation tasks for conditions and target texts.
  However, the encoding process of conditional texts lacks the understanding of target texts. 
  To this end, a spiral interaction architecture for encoder-decoder text diffusion~(\ModelName{}) is proposed.
  Concretely, the conditional information from encoder is designed to be captured by the diffusion decoder, while the target information from decoder is designed to be captured by the conditional encoder. 
  These two types of information flow run through multi-layer interaction spirally for deep fusion and understanding.
  \ModelName{} is evaluated on four text generation tasks, including paraphrase, text simplification, question generation, and open-domain dialogue generation. 
  Experimental results show that \ModelName{} achieves competitive performance among previous methods on all four tasks, demonstrating the effectiveness and generalization ability of the proposed method.\footnote{Code will be available at \url{https://github.com/lxchtan/DiffuSIA}}
\end{abstract}

\input{1-introduction}
\input{2-related}
\input{3-preliminary}
\input{4-method}
\input{5-experiments}

\vspace{-2mm}
\section{Conclusion}
\vspace{-1mm}
  In this paper, we have explored the encoder-decoder architecture for text diffusion, which offers greater flexibility due to its detachable encoder and decoder modules. 
  The flexibility of the model makes it extensible to multilingual and multimodal generation tasks for conditions and target texts.
  We proposed a spiral interaction architecture (\ModelName{}) that leverages the target information to improve the understanding of the conditional text. The results of our experiments show that \ModelName{} achieves competitive performance among previous methods on all four tasks, demonstrating the effectiveness and generalization ability of the proposed method. However, there is room for improvement in terms of dialogue generation tasks.  
  
\section*{Limitations}
While our model demonstrates good performance on various datasets, it falls short on tasks that demand higher natural language understanding capabilities, such as dialogue response generation. Improving natural language understanding will be a focus for future research. 
Additionally, our model incurs longer training times for improved performance, whereas pretrained-finetune workflow often require only 3-5 epochs of training to achieve better results on downstream tasks. 
Therefore, exploring ways to effectively utilize pre-trained language models is also an area of research we plan to investigate in the future.





\bibliography{custom}
\bibliographystyle{acl_natbib}


\end{document}

%% file: 1-introduction.tex
\section{Introduction}
  \begin{figure}[t]
  \centering
  \includegraphics[width=7.5cm]{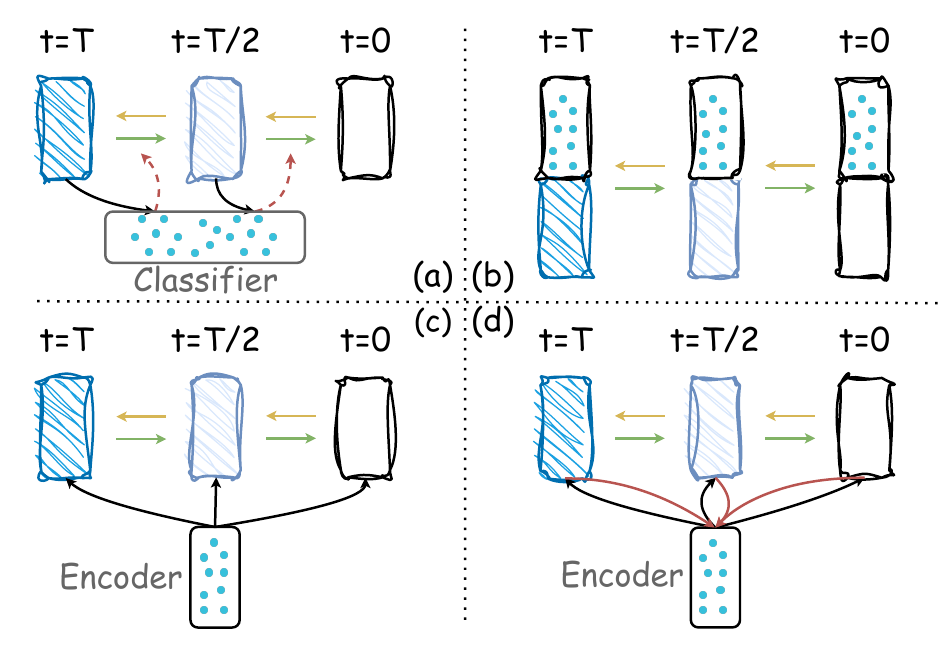}
  \caption{
    Comparison of existing methods for text diffusion.
    \textcolor{cyan}{Blue dots} denote conditional signals.
    (a) Diffusion-LM \cite{DBLP:journals/corr/abs-2205-14217}: classifier-guided text diffusion, where diffusion texts are sent to a pre-trained classifier and controlled by the return gradients.
    (b) DiffuSeq \cite{DBLP:journals/corr/abs-2210-08933}: single encoder-based text diffusion, where conditional text remains constant and is concatenated with target text during partially noising diffusion.
    (c) Encoder-decoder text diffusion: conditional text is encoded by a separate encoder and influences the generation process by cross-attention.
    (d) \ModelName: spiral interaction architecture for encoder-decoder text diffusion, where conditional text and target text perceived mutually through two splitting cross-attentions.
  }
  \vspace{-4mm}
  \label{fig: diffusion_ntypes}
  \end{figure}
  
  Diffusion models have recently become state-of-the-art for deep generative models, surpassing generative adversarial networks (GANs)~\cite{DBLP:conf/nips/GoodfellowPMXWOCB14} or normalizing flow~\cite{DBLP:conf/iclr/DinhSB17} in generative tasks such as image synthesis~\cite{DBLP:conf/nips/DhariwalN21, DBLP:conf/nips/HoJA20, DBLP:journals/corr/abs-2204-06125, DBLP:conf/cvpr/RombachBLEO22}. 
  Recently, different from traditional auto-regressive generation processing~\cite{radford2019language, DBLP:conf/acl/LewisLGGMLSZ20, DBLP:conf/acl/TanGTLXHGJ22}, the natural language processing community has also started to apply diffusion methods to the task of text generation considering their promising potentials~\cite{DBLP:conf/nips/AustinJHTB21, DBLP:journals/corr/abs-2205-14217, DBLP:journals/corr/abs-2208-04202, DBLP:journals/corr/abs-2210-08933}.
  Diffusion process typically operates in continuous space, which is naturally suitable for processing images.
  However, a major challenge to text diffusion lies in that text inherently operates in discrete space.

  Researchers have made efforts to applying diffusion models to various text generation tasks.
  For example, Diffusion-LM~\cite{DBLP:journals/corr/abs-2205-14217} designs an embedding step and a rounding step in the standard diffusion process~\cite{DBLP:conf/nips/HoJA20} for unconditional and controllable text generation.
  For conditional text generation, DiffuSeq~\cite{DBLP:journals/corr/abs-2210-08933} adopts partially noising processes with only a single Transformer encoder~\cite{DBLP:conf/nips/VaswaniSPUJGKP17} and is trained end-to-end in a classifier-free manner. 
  Despite the conditions can be integrated into the diffusion generation process, 
  the conditional encoder and the diffusion decoder are bound together and cannot be designed flexibly and independently. 
  Considering the limitations of the single Transformer encoder architecture for text diffusion, the encoder-decoder architecture shows its natural flexibility since two different modules can be designed for condition encoding and diffusion decoding respectively. 
  However, the sight of the other side of the coin should never be overlooked.
  This separation design makes the encoding of conditional text incapable of perceiving target text during the diffusion process, which might degrade the understanding of conditional text.
  But this issue has not been studied in previous work.

  Note that the generation process of diffusion is essentially non-autoregressive (NAR) with multiple iterations, thus the target information can be utilized to assist in understanding the conditional text without information leakage.
  In light of the above issues, a \textbf{s}piral \textbf{i}nteraction \textbf{a}rchitecture for encoder-decoder text \textbf{diffu}sion~(\ModelName) is proposed in this paper. 
  Comparison of existing methods for text diffusion are illustrated in Figure~\ref{fig: diffusion_ntypes}.
  The conditional information from encoder is designed to be captured by the diffusion decoder, while the target information from decoder is designed to be captured by the conditional encoder. 
  In detail, the encoder layer initially engages in interactions with the target text information through cross-attention to acquire the target-aware conditional (TaC) information. 
  Subsequently, the acquired TaC information is utilized in the interactions with the decoder layer through another cross-attention, deriving the condition-aware target (CaT) information. 
  These two types of information flow run through multi-layer interaction spirally, augmenting encoding and perception of both conditional text and target text.
  In this way, \ModelName{} is able to provide a flexible option for conditional text diffusion generation.
  Because of the NAR process of diffusion, the decoder does not require a causal mask.
  Besides, inspired by previous works~\cite{DBLP:journals/corr/abs-2208-04202, DBLP:journals/corr/abs-2211-04236, DBLP:journals/corr/abs-2211-15089}, the diffusion generation result from previous timesteps are used for self-conditioning~\cite{DBLP:journals/corr/abs-2208-04202} to predict the target at the current timestep.

  To measure the effectiveness of the proposed method, following the setting of \citet{DBLP:journals/corr/abs-2210-08933}, we evaluate the performance on four popular text generation tasks, including paraphrase, text simplification, question generation, and open-domain dialogue generation. 
  Experiments on these text generation tasks show that our method achieves competitive performance.
  These results verify the effectiveness of the spiral interactions for encoder-decoder text diffusion, and the generalization ability over various text generation tasks.
  To facilitate others to reproduce our results, we will publish all source code later.
  
  In summary, our contributions in this paper are three-fold:
  1) This paper makes the exploration of applying the encoder-decoder architecture for text diffusion.
  2) A spiral interaction architecture is proposed for encoder-decoder text diffusion, which is composed of the target-aware conditional (TaC) and condition-aware target (CaT) information flows.
  3) Experiments on four types of text generation tasks verify the effectiveness and generalization ability of the proposed method.

%% file: 2-related.tex
\section{Related Work}

  In recent years, diffusion models have achieved great success in the domain of image synthesis~\cite{DBLP:conf/icml/NicholDRSMMSC22, DBLP:journals/corr/abs-2204-06125, DBLP:journals/corr/abs-2209-15264, DBLP:conf/cvpr/RombachBLEO22}. 
  Because of its amazing generation quality, some works apply diffusion model to the domain of text generation. 
  There are two general lines of work on text diffusion, namely discrete diffusion on discrete data~\cite{DBLP:conf/nips/HoogeboomNJFW21, DBLP:conf/nips/AustinJHTB21, DBLP:conf/iclr/SavinovCBEO22, DBLP:journals/corr/abs-2210-16886, DBLP:journals/corr/abs-2211-15029} and continuous diffusion on discrete data. 
  In this paper, we study the latter.

  \paragraph{Unconditional and Controllable Text Diffusion} Bit Diffusion~\cite{DBLP:journals/corr/abs-2208-04202} uses real numbers to model the bits of data for enabling continuous state diffusion models to generate discrete data. Besides, \textit{self-conditioning} and \textit{asymmetric time intervals} that greatly improve the sample quality.
  Diffusion-LM~\cite{DBLP:journals/corr/abs-2205-14217} maps discrete tokens into continuous latent variable by adding an embedding step and a rounding step to the standard diffusion process with designing a training objective to learn the embedding.
  It achieves more complex controllable text generation through continuous diffusion. 

  \paragraph{Conditional Text Diffusion} 
  DiffuSeq~\cite{DBLP:journals/corr/abs-2210-08933} adopts partially noising processes with only a single Transformer encoder and trained end-to-end in a classifier-free manner to extend Diffusion-LM for sequence-to-sequence (Seq2Seq) generation tasks. 
  Considering the importance of embedding space for the diffusion process, SED~\cite{DBLP:journals/corr/abs-2211-04236} uses a BERT to generate embeddings for diffusion input tokens, with the training objective of Diffusion-LM and self-conditioning skill from Bit Diffusion. Besides, classifier-free guidance~\cite{DBLP:journals/corr/abs-2207-12598} are performed to allows leveraging both the unconditional and conditional abilities of a model to improve its conditional generations.
  CDCD~\cite{DBLP:journals/corr/abs-2211-15089} is a framework for continuous diffusion models of categorical data with score interpolation and time warping based on score matching diffusion models~\cite{DBLP:conf/nips/SongE19, DBLP:conf/iclr/0011SKKEP21}. It adopts an encoder-decoder (ED) architecture for machine translation.
  The potential of applying ED architectures to more diffusion text generation tasks still needs to be explored.
  It should be noted that a concurrent study SeqDiffuSeq~\cite{DBLP:journals/corr/abs-2212-10325} also studies applying encoder-decoder for text diffusion.
  SeqDiffuSeq extends the continuous text diffusion model to sequence-to-sequence text generation under the encoder-decoder architecture. Two techniques of self-conditioned denoising and token-level adaptive noise schedule are also adopted in SeqDiffuSeq.
  
  Compared with SeqDiffuSeq, we analyze the defects of the ED architecture and further investigate the effect of different numbers of encoder and decoder layers to text diffusion. 
  To the best of our knowledge, this paper makes the first attempt to mitigate the issue of conditional text not perceiving target text when applying encoder-decoder for conditional text generation with diffusion. 
  Additionally, \ModelName{} is proposed to strengthen interactions between conditional text and target text.

%% file: 3-preliminary.tex
\section{Preliminaries}
  \paragraph{Unconditional Diffusion}
  Diffusion models involve perturbing data with increasing levels of random noise, then removing the noise to generate new samples. 
  This process is known as diffusion, and is the key element of three main formulations of diffusion models, i.e., denoising diffusion probabilistic models (DDPMs)~\cite{DBLP:conf/nips/HoJA20, DBLP:conf/iclr/SongME21}, score-based generative models (SGMs)~\cite{DBLP:conf/nips/SongE19, DBLP:conf/nips/0011E20}, and stochastic differential equations (Score SDEs)~\cite{DBLP:journals/corr/abs-2206-00364, DBLP:conf/nips/SongDME21, DBLP:journals/corr/abs-2208-09141}. 
  In this work, we study DDPMs.

  Formally, given a data distribution $\bx_0\sim q(\bx_0)$, the forward Markov process generates a sequence of random variables $\bx_1, \bx_2, ..., \bx_T$ with transition kernel $q(\bx_t|\bx_{t-1}) = \mathcal{N}(\bx_t;\sqrt{1-\beta_t}\bx_{t-1}, \beta_t\boldsymbol{I})$, where $\beta_t \in (0, 1)$ is a hyperparameter chosen ahead of model training as different variance scales. The final state $\bx_T$ is almost Gaussian in distribution, so we have $q(\bx_T) \approx \mathcal{N}(\bx_T; \boldsymbol{0}, \boldsymbol{I})$. 
  For the reverse Markov process, a learnable reverse transition kernel $ p_{\theta}(\bx_{t-1}|\bx_{t}) = \mathcal{N}(\bx_{t-1}; \mu_{\theta}(\bx_t, t), \varSigma_{\theta} (\bx_t, t))$ is trained to fit the posterior distribution $q(\bx_{t-1}|\bx_t,\bx_0) =  \mathcal{N}(\bx_{t-1}; \tilde{\boldsymbol{\mu}}_t(\bx_t, \bx_0), \tilde\beta_t \boldsymbol{I})$ where $\tilde\bmu_t(\bx_t, \bx_0) \defeq \frac{\sqrt{\bar\alpha_{t-1}}\beta_t }{1-\bar\alpha_t}\bx_0 + \frac{\sqrt{\alpha_t}(1- \bar\alpha_{t-1})}{1-\bar\alpha_t} \bx_t$ and $\tilde\beta_t \defeq \frac{1-\bar\alpha_{t-1}}{1-\bar\alpha_t}\beta_t$ with the notation $\alpha_t \defeq 1-\beta_t$ and $\bar\alpha_t \defeq \prod_{s=1}^t \alpha_s$.
  The training objective can be simplified as: 
  \begin{equation} \label{Lsimple}
    \small
    \Lsimple (\cx{0}) = \sum_{t=1}^T \mathop{\mathbb{E}}_{q(\cx{t} \mid \cx{0})} || \mu_\theta(\cx{t}, t) - \tilde\bmu_t(\bx_t, \bx_0) ||^2.
  \end{equation}
  Once the forward process is completed, the reverse denoising process is tasked to gradually reconstruct the original data $\mathbf{x}_0$ via sampling from $\mathbf{x}_T$ by learning a diffusion model. 

  \paragraph{Continuous Diffusion on Embedding Space}
  Diffusion-LM~\cite{DBLP:journals/corr/abs-2205-14217} proposes continuous diffusion on the embedding space for text generation. 
  In the forward process, an \textit{embedding step} is designed to introduce a Markov transition from discrete words $\ww$ to $\cx{0}$ that is parametrized by $\qphi(\cx{0} | \ww) = \mathcal{N} (\Emb(\ww), \sigma_0 I )$. 
  In the reverse process, a trainable \textit{rounding step} is added and parametrized by $\ptheta(\ww \mid \cx{0}) = \prod_{i=1}^n \ptheta(w_i \mid x_i)$, where $\ptheta(w_i \mid x_i)$ is a $\softmax$ distribution. Based on Eq.~(\ref{Lsimple}), the training objective is modified as: 
  \begin{equation}
    \begin{split}
      & \mathcal{L}^\text{e2e}_\text{simple} (\ww) = \mathop{\mathbb{E}}_{\qphi(\cx{0:T} | \ww)}  [\Lsimple(\cx{0}) + \\ & || \Emb(\ww) - \mu_\theta(\cx{1}, 1) ||^2  - \log \ptheta(\ww | \cx{0}) ].
    \end{split}
  \end{equation}

  \paragraph{Classifier-free Guidance}
  Extending the guidance method proposed by \citet{DBLP:conf/nips/DhariwalN21}, \textit{semantic diffusion guidance} (SDG)~\cite{DBLP:journals/corr/abs-2112-05744} allows fine-grained and continuous control of model class, including either language or image guidance, or both.
  Furthermore, a classifier-free guidance method is proposed that is more effective at controlling generation~\cite{DBLP:journals/corr/abs-2207-12598, DBLP:journals/corr/abs-2204-06125}. 
  Let unconditional denoising diffusion model $p_\theta(\mathbf{x})$ be parameterized through a score estimator $\boldsymbol{\epsilon}_\theta(\mathbf{x}_t, t)$ and the conditional model $p_\theta(\mathbf{x} \vert c)$ be parameterized through $\boldsymbol{\epsilon}_\theta(\mathbf{x}_t, t, c)$. These two models can be learned via a single neural network. 
  Precisely, a conditional diffusion model $p_\theta(\mathbf{x} \vert c)$ is trained on paired data $(\mathbf{x}, c)$, where the conditioning information $c$ is discarded periodically and randomly, so that the model knows how to generate unconditionally as well, i.e. $\boldsymbol{\epsilon}_\theta(\mathbf{x}_t, t) = \boldsymbol{\epsilon}_\theta(\mathbf{x}_t, t, c=\varnothing)$. 

  In this paper, we focus on the sequence-to-sequence text generation tasks which  produce a target sequence $\ww^x = \{w^x_1, ..., w^x_n\}$ conditioning on the source sequence $\ww^c = \{w^c_1, ..., w^c_m\}$. Different from \citet{DBLP:journals/corr/abs-2207-12598}, conditional information is involved all the time and not discarded, which has been proved effective in \citet{DBLP:journals/corr/abs-2210-08933}. Thus the training objective becomes:
  \begin{equation}
    \begin{split}
    & \mathcal{L}_{\text{VLB}}  = \mathop{\mathbb{E}}_{\qphi(\cx{0:T} | \ww, \cc)}[
      \sum_{t=2}^T||\cx{0}-f_{\theta}(\cx{t}, \cc, t)||^2 + \\ & ||\textsc{Emb}(\ww^x)-f_{\theta}(\cx{1}, \cc, 1)||^2-\log p_{\theta}(\ww^{x}|\cx{0})].
    \end{split}
  \end{equation}

%% file: 4-method.tex
\section{Approach}

  In this section, we first describe the encoder-decoder architecture for encoding the conditional text.
  To augment encoding and perception of both conditional text and target text, a spiral interaction modification is then proposed.
  Finally, we briefly introduce the technique of \textit{self-conditioning}~\cite{DBLP:journals/corr/abs-2208-04202} adopted in our method. 

  \subsection{Encoder-Decoder Diffusion} \label{sec: DiffuED}
  This paper refers to the component that encodes the conditional text as the encoder, and that denoises the target text as the decoder.

  \paragraph{Conditional Encoder (CE)} 
  To encode conditional text, an embedding function is used to map conditional tokens to hidden states, i.e., $\cc^0 = \Emb_c{(\ww^c)}$.
  The output of a conditional encoder layer is used as the input of the next layer.
  Readers can refer to \citet{DBLP:conf/nips/VaswaniSPUJGKP17} for details of Transformer encoder. 
  Formally, the calculation at the \emph{m}-th encoder layer is denoted as:
  \begin{align} \label{eq: encoder}
    \cc^{m+1} = \mathop{\TransformerEncoder}( \cc^m ),
  \end{align}
  where $m \in \{0, ..., L_e - 1\}$ and $L_e$ denotes the number of Transformer encoder layers. 
  $\cc^m \in \mathbb{R}^{k_c \times d_c}$, where $k_c$ denotes the length of conditional text and $d_c$ denotes the dimension of conditional text embedding vectors.

  \paragraph{Target Decoder (TD)} 
  To map target tokens to continuous representations, another embedding function is adopted, i.e., $\xx^0 = \Emb_x{(\ww^x)}$.
  Then, a Transformer decoder layer~\cite{DBLP:conf/nips/VaswaniSPUJGKP17} is used as the input of the next layer. 
  Formally, the calculation at the \emph{n}-th decoder layer is denoted as:
  \begin{equation}
    \xx^{n+1} = \mathop{\TransformerDecoder}( \xx^n, \cc^{L_e}), \label{eq:TD}
  \end{equation}
  where $n \in \{0, ..., L_d - 1\}$ and $L_d$ denotes the number of Transformer decoder layers. 
  $\xx^l \in \mathbb{R}^{k_x \times d_x}$, where $k_x$ denotes the length of conditional text and $d_x$ denotes the dimension of target text embedding vectors.
  The representations of conditional text from the last encoder layer $\cc^{L_e}$ is fused into the target representation to control the generation process by cross-attention mechanism as:
  \begin{equation}
    \small
    \mathop{\CA}({\xx^n}\boldsymbol{W}_q^n, \cc^{L_e}\boldsymbol{W}^n_k, \cc^{L_e}\boldsymbol{W}^n_v),
  \end{equation}
  where $\boldsymbol{W}^n_q\in \mathbb{R}^{d_x \times d_x}$ and $\boldsymbol{W}^n_{\{k, v\}} \in \mathbb{R}^{d_c \times d_x}$.
  Different from the regular Transformer decoder, the causal mask is not necessary, as the generation process of diffusion is non-autoregressive (NAR).

  It is notable that only one time of encoding of the conditional text is required here, since $\cc^{L_e}$ is independent of timestep $t$, which is computation-efficient.
  However, the lack of information involving $\xx^t$ degrades the representation capability of $\cc^{L_e}$, compared with the full self-attention operation in DiffuSeq.
  Thus, a spiral interaction architecture is introduced next to addressee this issue.

  \subsection{Spiral Interaction Architecture} \label{sec: SIA}
    \begin{figure}[t]
      \centering
      \includegraphics[width=7.5cm]{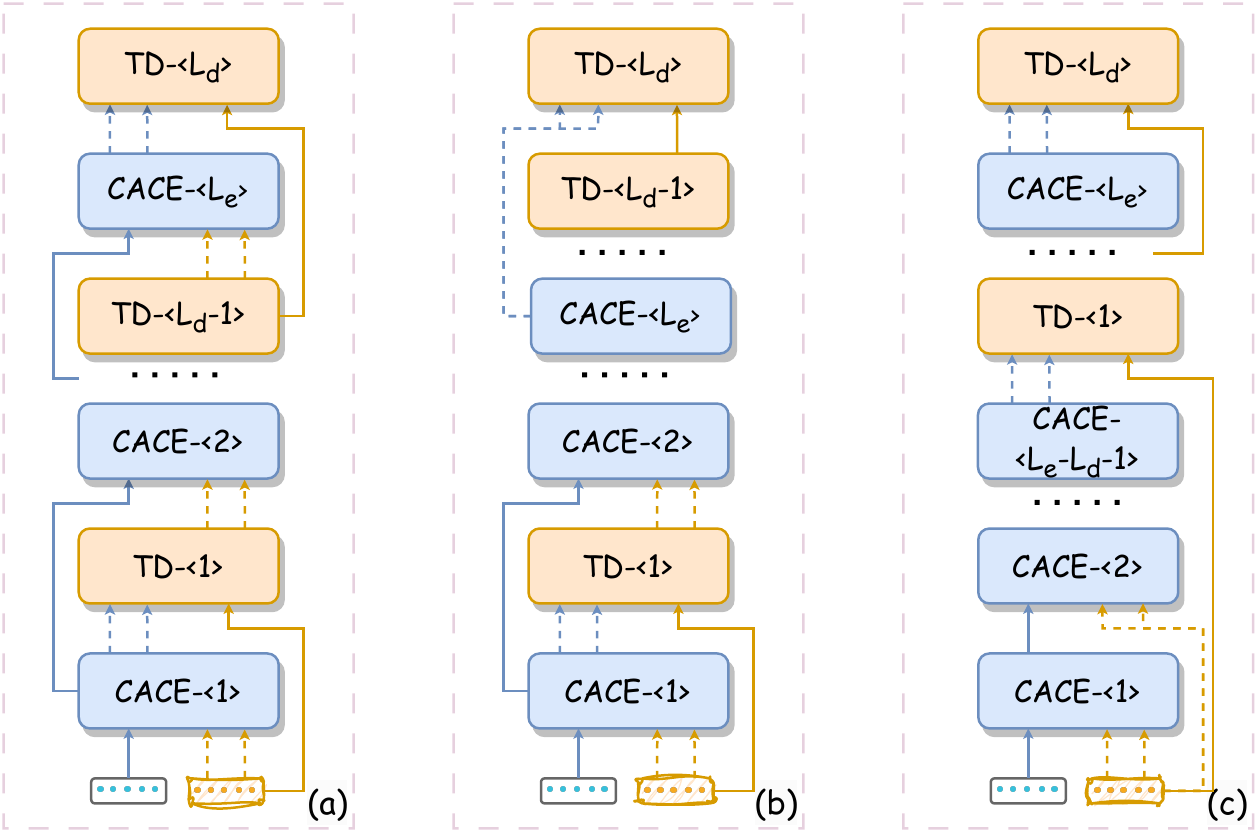}
      \caption{Illustration of spiral interaction architecture (SIA). The sub-figures of (a), (b), (c) show the cases of $L_e=L_d$, $L_e < L_d$ and $L_e > L_d$ respectively. \textcolor[RGB]{108, 142, 191}{Blue} represents the TaC flow and \textcolor[RGB]{215, 155, 0}{yellow} represents the CaT flow. Solid lines denote query, while dashed lines denote key and value in cross-attention.} 
      \vspace{-4mm}
      \label{fig: SIA}
    \end{figure}
    
  To augment encoding and perception of both conditional text and target text, these two information flows are designed to be spirally intertwined.
  An overview of the proposed spiral interaction architecture for encoder-decoder text diffusion is illustrated in Figure~\ref{fig: SIA}.

  \paragraph{Conditional Encoder with Cross-Attention (CACE)} 
  Cross-attention mechanism is introduced here to let the conditional information attend to the target information. 
  Then, Eq.~(\ref{eq: encoder}) is modified as:
  \begin{equation} \label{eq: CACE2}
    \cc^{m+1}_t = \mathop{\CACE}( \cc^m_t, \xx_t^0 ).
  \end{equation}
  Furthermore, \ModelName{} has no concern of information leakage due to its NAR process.
  Correspondingly, Eq.~(\ref{eq:TD}) is modified as:
  \begin{equation} \label{eq: TD2}
    \xx^{n+1}_t = \mathop{\TransformerDecoder}( \xx^n_t, \cc^{L_e}_t).
  \end{equation}

  \paragraph{Splitting and Interweaving} 
  In order to further strengthen the information interaction between conditions and targets, a strategy of splitting and interweaving is designed.
  As shown in Figure~\ref{fig: SIA}, the layers of CACE and TD are split and interleaved to form spiral interactions. 
  The encoder layers of CACE involve in interactions with the target text information through cross-attention to acquire the target-aware conditional (TaC) representations. 
  Thus Eq.~(\ref{eq: CACE2}) is modified as:
  \begin{equation}
    \cc^{m+1}_t = \mathop{\CACE}( \cc^m_t, \xx_t^{n} ).
  \end{equation}
  
  Subsequently, the acquired TaC information is utilized in the interactions with the decoder layers of TD through cross-attention, deriving the condition-aware target (CaT) representations. Thus Eq.~(\ref{eq: TD2}) is modified as:
  \begin{equation}
    \xx^{n+1}_t = \mathop{\TransformerDecoder}( \xx^n_t, \cc^{m+1}_t).
  \end{equation}
  
  We consider three cases to accommodate the interactions of CACE and TD with different number of layers as:
  \begin{itemize}
    \item $L_e = L_d$. The encoding process is accomplished by simply interleaving CACE with each layer of TD in this setup.
    \item $L_e < L_d$. The interleaving process operates from layer $0$ to $L_e - 1$. After that, individual diffusion decoding with Eq.~(\ref{eq: TD2}) is conducted. 
    \item $L_e > L_d$. The individual conditional encoding with Eq.~(\ref{eq: CACE2}) is first conducted from layer $0$ to $L_e - L_d - 1$. After that, the interleaving process is conducted.
  \end{itemize}
  These three cases provide corresponding strategies for models in various situations.

  \subsection{Self-Conditioning} \label{sec: SC}

  In the reverse process, the denoising function $f_{\theta}(\cx{t}, \cc, t)$ is only conditioned on the previous updated noisy samples $\cx{t}$, not directly on the function prediction $\cx{0}^t = f_{\theta}(\cx{t+1}, \cc, t+1)$, discarding the information of predictions from the previous step. 
  \textit{Self-conditioning}~\cite{DBLP:journals/corr/abs-2208-04202} is proposed to address the issue by taking $\cx{0}^t$ into account with a modification denoising function as:
  \begin{equation} \label{eq: self-cond}
    \cx{0}^{t-1} = f_{\theta}(\cx{t}, \cx{0}^t, \cc, t).
  \end{equation}
  Providing the model with direct access to the predictions it produced in the previous sampling step allows for a more efficient utilization of its capacity. In this way it can refine previous predictions, instead of constructing them from scratch in each step.~\cite{DBLP:journals/corr/abs-2208-04202, DBLP:journals/corr/abs-2211-15089, DBLP:journals/corr/abs-2211-04236}

  Following the setting in \citet{DBLP:journals/corr/abs-2208-04202}, with 50\% probability, we set $f_{\theta}(\cx{t}, \cx{0}^t=\boldsymbol{0}, \cc, t)$ which falls back to modeling without \textit{self-conditioning}. No back-propagating through the first estimated $\cx{0}^t$, the increase of additional training time is less than 25\%.
  In practice, to approximate the inference behavior at train time while remaining computationally efficient, the first estimated $\cx{0}^t$ is calculated as $\bar{\mathbf{x}}_{0}^t = f_{\theta}(\cx{t}, \boldsymbol{0}, \cc, t)$. Then we perform a second forward pass using a stop gradient to obtain $\cx{0}^{t-1} = f_{\theta}(\cx{t}, \bar{\mathbf{x}}_{0}^t, \cc, t)$.
  At inference time, we always estimate $\cx{0}$ based on Eq.~(\ref{eq: self-cond}).
  To combine the information of previous estimation, there are two simple method can be tried. The first one is that we concatenate $\cx{0}^{t-1}$ and $\cx{0}^{t}$ through the hidden dimension with a linear projection, while another one is that we directly add them together. 
  The experiment results show that the first one is more powerful.

%% file: 5-experiments.tex
\section{Experiments}
    \begin{table}[t]
      \centering
      \setlength{\tabcolsep}{3.0pt}
      \resizebox{.95\columnwidth}{!}{
      \begin{tabular}{lccccc}
      \toprule
        Tasks & CLens & TLens & BS & Steps/$k$ & T/$h$  \\ 
      \midrule
        QQP (Paraphrase) & 32 & 32 & 1400 & 50 & 12    \\
        Wiki-Auto (TS) & 128 & 64 & 2048 & 80 & 35       \\
        Quasar-T (QG) & 64 & 32 & 1400 & 50 & 12    \\
        CCD (DG) & 64 & 64 & 2048 & 140 & 45       \\
      \bottomrule
      \end{tabular}
      }
      \caption{
        Detail settings for four different tasks. 
        CLens means maximum length of conditional text.
        TLens means maximum length of target text.
        BS means batch size.
        Steps means learning steps.
        T means approximate training time of \ModelName{} on 4x A100 GPUs.
      }
      \vspace{-4mm}
      \label{tab: Settings}
    \end{table}
    
    \begin{table*}[t]
      \centering
      \setlength{\tabcolsep}{3.0pt}
      \resizebox{1.0\linewidth}{!}{
      \begin{tabular}{l|ccc|ccc|ccc|ccc}
      \toprule
        \multirow{2}{*}{\backslashbox{Models}{Metrics}}  &  \multicolumn{3}{c|}{QQP (Paraphrase)}  & \multicolumn{3}{c|}{Wiki-Auto (TS)} &  \multicolumn{3}{c|}{Quasar-T (QG)}    &  \multicolumn{3}{c}{CCD (DG)} \\ &  
                                                                  BLEU &  ROUGE$_L$ &  BERTS & 
                                                                  BLEU &  ROUGE$_L$ &  BERTS &  BLEU &  ROUGE$_L$ &  BERTS &  BLEU &  ROUGE$_L$ &  BERTS  \\
      \midrule
        Transformer \cite{DBLP:conf/nips/VaswaniSPUJGKP17}       & 5.80 & 24.89 & 53.92 & 24.45 & 50.58 & 75.90 & 3.64 & 19.94 & 53.34 & \textbf{1.89} & 10.39 & 47.81    \\
        GPT2-Large~\cite{radford2019language} & 20.59 & 54.15 & 83.63 & 26.93 & 51.11 & 78.82 & 11.10 & 32.15 & \textbf{63.46} & 1.25 & 10.02 & \textbf{52.93}  \\
        GPVAE~\cite{DBLP:journals/corr/abs-2204-01227} & 24.09 & 58.86 & \textbf{84.66} & 33.92 & 58.28 & 81.66 & 12.51 & 33.90 & 63.08 & 1.10 & 10.09 & 43.17       \\
        LevT (NAR) \cite{DBLP:conf/nips/GuWZ19}  & 22.68 & 57.95 & 83.44 & 20.52 & 44.02 & 72.54 & 9.30 & 28.93 & 54.91 & 1.58 & 5.50 & 47.60       \\
        DiffuSeq \cite{DBLP:journals/corr/abs-2210-08933} & 24.13 & 58.80 & 83.65 & 36.22 & 58.49 & 81.26 & \textbf{17.31} & \textbf{36.65} & 61.23 & 1.39 & \textbf{10.56} & 51.31      \\
      \midrule
        DiffuSIA    & \textbf{24.95} & \textbf{59.55} & 83.62 & \textbf{37.03} & \textbf{59.63} & \textbf{81.90} & 17.12 & 35.13 & {62.19} & 1.13 & 9.61 & 50.58    \\ 
      \bottomrule
      \end{tabular}
      }
      \caption{Evaluation results on four test sets in terms of automated evaluation. The results of baselines is copied from \citet{DBLP:journals/corr/abs-2210-08933}.
      Numbers in \textbf{bold} denoted that the best score. BERTS is the short of BERTScore.}
      \label{tab: results}
    \end{table*}
  
    \subsection{Datasets}
    Following \citet{DBLP:journals/corr/abs-2210-08933}, experiments on four different sequence-to-sequence text generation tasks were conducted to validate the effectiveness of the proposed \ModelName{}: 

    \vspace{-2mm}
    \paragraph{Paraphrase} The {Quora Question Pairs} (QQP) dataset\footnote{\url{https://www.kaggle.com/c/quora-question-pairs}}, extracted from the question-answering forum Quora, is used for paraphrase evaluation, where the positive question pairs are used to evaluate models' ability to generate a restatement of a question expressing the same meaning.

    \vspace{-2mm}
    \paragraph{Text Simplification (TS)} The {Wiki-Auto} dataset~\cite{DBLP:conf/acl/JiangMLZX20} is a text simplification dataset, consisting of 666K complex-simple sentence pairs with revision alignment, which is used to revise complex text with simplified grammar and word choice.

    \vspace{-2mm}
    \paragraph{Question Generation (QG)}  The {Quasar-T} dataset~\cite{DBLP:journals/corr/DhingraMC17} is used for evaluating question generation which aims to generate related questions with a given context. The preprocessed data of \citet{DBLP:conf/acl/SunLLJ18} is used following \citet{DBLP:journals/corr/abs-2210-08933}.

    \vspace{-2mm}
    \paragraph{Open Domain Dialogue (DG)} The {Commonsense Conversation Dataset} (CCD)~\cite{DBLP:conf/ijcai/ZhouYHZXZ18} extracted from single-round dialogue in Reddit, is used for evaluating open-domain dialogue, the task of generating informative feedback based on the dialogue context.
    
    \subsection{Baselines}
    The following methods were considered as baselines: 
    \textbf{(1) Transformer} \cite{DBLP:conf/nips/VaswaniSPUJGKP17} is an encoder-decoder architecture that performs text generation in an autoregressive (AR) manner.
    \textbf{(2) GPT-2} \cite{radford2019language} is a uni-directional pre-trained language model as a strong AR baseline. 
    \textbf{(3) GPVAE} \cite{DBLP:journals/corr/abs-2204-01227} augments a pre-trained T5~\cite{DBLP:journals/jmlr/RaffelSRLNMZLL20} with variational attention~\cite{DBLP:conf/coling/BahuleyanMVP18,DBLP:conf/nips/DengKCGR18, DBLP:conf/ijcai/Wang019b} to improve the generation diversity.
    \textbf{(4) LevT} \cite{DBLP:conf/nips/GuWZ19} is a partially autoregressive model devised for more flexible and amenable sequence generation, chosen as a conventional NAR baseline. 
    \textbf{(5) DiffuSeq} \cite{DBLP:journals/corr/abs-2210-08933} uses an encoder-only Transformers architecture and partially noising to adapt text diffusion model to sequence-to-sequence task.

    \subsection{Implementation Details}
    
    For a fair comparison with DiffuSeq consisting of a single Encoder with 12 layers, our \ModelName{} was based on the six to six layers encoder-decoder Transformer~\cite{DBLP:conf/nips/VaswaniSPUJGKP17}. 
    The encoder embedding dimension is set to $768$, while the decoder embedding dimension is set to $128$.
    Each encoder/decoder layer was under the setting of \emph{bert-base-uncased}.
    The diffusion timestep information was formulated as timestep embedding which was added to the word embedding. 

    The diffusion steps was set to $2000$, and the initial noise schedule was set to \textit{sqrt}.
    Schedule sampler was set to \textit{lossaware} as \citet{DBLP:journals/corr/abs-2210-08933}.
    The AdamW method~\cite{DBLP:conf/iclr/LoshchilovH19} was employed for optimization. 
    The learning rate was initialized as $1e\text{-}4$ and was decayed linearly down to $0$. 
    As shown in Table~\ref{tab: Settings}, for different tasks, different batch size, learning steps and maximum utterance length were set.
    The strategy of Maximum Bayes Risk (MBR)~\cite{DBLP:conf/naacl/KumarB04} with the size of candidate samples $|\mathcal{S}|=10$ was performed for decoding. 
    All experiments were run on four NVIDIA Tesla A100 80G GPUs. 
    Half-precision floating-point format $FP16$ was applied to accelerate training and decoding process.
    All code was implemented in the PyTorch framework\footnote{\url{https://pytorch.org/}}.
  
  \subsection{Metrics}
    To evaluate the quality of the generated text, we employed the standard string-similarity-based metrics BLEU~\citep{DBLP:conf/acl/PapineniRWZ02} and ROUGE~\citep{lin2004rouge}. 
    Besides, BERTScore~\citep{DBLP:conf/iclr/ZhangKWWA20} was also employed to help measure the semantic similarity between the generated sentences and the references. Higher is better for all metrics.\footnote{The evaluation codes are provided by \citet{DBLP:journals/corr/abs-2210-08933}.}
  
  \subsection{Evaluation Results}
    Table~\ref{tab: results} presents the evaluation results of \ModelName{} and previous methods on the four test sets. 
    Our proposed \ModelName{} achieved competitive performance over these baseline methods on Wiki-Auto and QQP, outperformed conventional generation methods (except DiffuSeq) on Quasar-T, but the performance was not as good as those on CCD.
    In particular, \ModelName{} outperformed the best performing baseline by large margins of 0.86\% BLEU and 0.69\% ROUGE$_L$, but left behind 1.04\% BERTScore on QQP. 
    In terms of Wiki-Auto, \ModelName{} outperformed the best performing baseline by large margins of 0.81\% BLEU, 1.14\% ROUGE$_L$ and 0.24\% BERTScore respectively.
    In terms of Quasar-T, \ModelName{} outperformed the best performing conventional baseline by large margins of 4.61\% BLEU and 1.23\% ROUGE$_L$, but left behind 1.27\% BERTScore. 
    Compared with DiffuSeq, \ModelName{} outperformed it by 0.96\% BERTScore, but left behind 0.19\% BLEU and 1.52\% ROUGE$_L$ on Quasar-T.
    In terms of CCD, the performance of \ModelName{} was not as good as the baselines.
    Compared with the other three tasks, DG task required deeper natural language understanding and reasoning abilities.
    From these results, it can be seen that there is still room for further improvement.

  \begin{table}[t]
    \centering
    \setlength{\tabcolsep}{3.0pt}
    \resizebox{.9\columnwidth}{!}{
    \begin{tabular}{lccc}
    \toprule
    {Models} & BLEU &  ROUGE$_L$ &  BERTScore  \\ 
    \midrule
      \ModelName{} & 24.95 & 59.55 & 83.62 \\
      \quad w/o. SI & 24.48 & 59.00 & 83.30 \\ 
      \quad w/. A-Type SC & 23.85 & 58.99 & 82.87 \\ 
      \quad w/o. SC & 23.68 & 58.24 & 82.71 \\ 
    \midrule
      DiffuED & 24.26 & 59.18 & 83.92 \\
      \quad w/. A-Type SC & 24.44 & 59.29 & 83.43 \\ 
      \quad w/o. SC & 23.77 & 58.44 & 83.01 \\ 
      \hdashline
      \quad PreEnc S-BERT & 23.19 & 58.33 & 83.36 \\
      \quad PreEnc T-BERT & 24.18 & 58.75 & 83.54 \\
    \bottomrule
    \end{tabular}
    }
    \caption{
      Experiments of the modified architecture on QQP.
      SI indicates Splitting and Interweaving.
      SC indicates Self-Conditioning.
      A-Type SC indicates self-conditioning is directly add to $\xx_t$ as described in Sec.~\ref{sec: SC}.
      DiffuED is the pure encoder-decoder diffusion as described in Sec.~\ref{sec: DiffuED}.
      PreEnc indicates using pretrained encoder.
      S-BERT is the short of Sentence-BERT.
      T-BERT is the short of tinyBERT.
    }
    \vspace{-4mm}
    \label{tab: ablations}
  \end{table}

  \subsection{Ablation Study}
    To further verify the effectiveness of the proposed \ModelName{}, comparison with the encoder-decoder diffusion architecture described in Sec.~\ref{sec: DiffuED}, namely DiffuED, was conducted on the QQP dataset.
    As demonstrated in Table~\ref{tab: ablations}, \ModelName{} outperformed DiffuED by margins of 0.69\% BLEU, 0.37\% ROUGE$_L$, illustrating the effectiveness of the interweaved TaC and CaT flows.
    Besides, ablating the technique of splitting and interweaving~(SI) resulted in degraded performance on all three metrics, indicating that the spiral architecture was crucial for modeling the interactions between conditional text and target text.

    On the other hand, self-conditioning~(SC) was ablated, denoted as \ModelName{} {w/o. SC}, to explore its effect on models.
    The performance of both models decreases after removing the SC, illustrating the importance of the SC.
    Besides, self-conditioning was also directly added to the inputs of decoder, denoted as \ModelName{} {w/. A-Type SC}, to compare with the concatenated-type using in \ModelName{}, denoted as C-Type SC.
    It can be seen that \ModelName{} outperformed \ModelName{} {w/. A-Type SC}, but no performance degradation for DiffuED {w/. A-Type SC}.
    The results indicated that C-Type SC is more robust than the A-Type SC.

  \subsection{Analysis}
\begin{figure}[t]
    \centering
    \includegraphics[width=7.5cm]{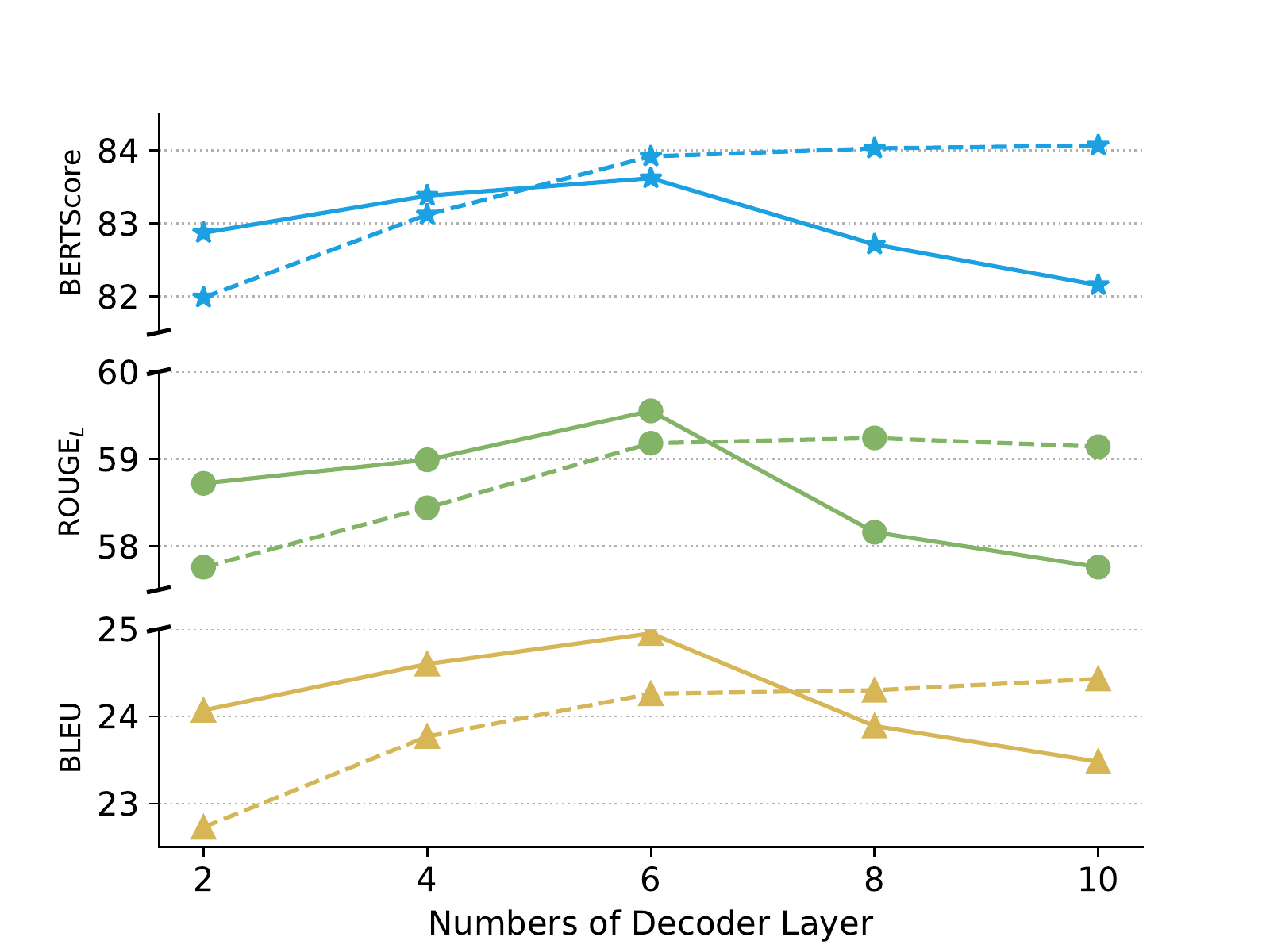}
    \caption{
      Impact of different numbers of decoder layers to \ModelName{} and DiffuED on the test set of QQP dataset.
      Solid lines for \ModelName{}, and dashed lines for DiffuED. 
    }
    \vspace{-4mm}
    \label{fig: trend_DL}
  \end{figure}

    \vspace{-1mm}
    \paragraph{Impact of the number of encoder and decoder layers.}
      We explored how the number of encoder and decoder layers affected the performance of \ModelName{}.
      To ensure a fair comparison, the number of encoder layers $L_e$ and the number of decoder layers $L_d$ were under the restraint of $L_e + L_d = 12$. 
      For DiffuED, it's easy to set different values for encoder and decoder. 
      For \ModelName{}, the architecture shown in Figure~\ref{fig: SIA} was applied.
      As shown in Figure~\ref{fig: trend_DL}, \ModelName{} showed different trend from that of DiffuED.
      As the number of decoder layers increased (meanwhile the number of encoder layers decreased), the performance of DiffuED was consistently improved on the QQP dataset.
      On the other hand, the performance of \ModelName{} was improved, as the number of decoder layers increased at the beginning.
      $L_d=6$ achieved the best performance.
      After that, the performance of \ModelName{} dropped as the number of decoder layers further increased.
      These results indicated that the spiral interaction architecture showed best performance under a symmetrical structure of encoder and decoder.

    \vspace{-1mm}
    \paragraph{Pre-trained Encoder.}
      Experiments of exploring different pre-trained language models for diffusion generation process were conducted.
      The encoder of DiffuED was initialized using a pre-trained 6-layer BERT model.
      Specifically, DiffuED PreEnc S-Bert was initialized using Sentence-BERT~\cite{reimers-2019-sentence-bert}\footnote{\url{https://huggingface.co/sentence-transformers/paraphrase-TinyBERT-L6-v2}}, while DiffuED PreEnc T-Bert was initialized using tinyBERT~\cite{DBLP:conf/emnlp/JiaoYSJCL0L20}\footnote{\url{https://huggingface.co/huawei-noah/TinyBERT_General_6L_768D}}.
      The results were shown in the last two rows in Table~\ref{tab: ablations}.
      The pre-trained models had instead played a negative role, for the gap between conditional text encoder and diffusion process decoder.
      It suggested that further improvements are needed in effectively utilizing pre-trained language models.

  \begin{table}[t]
    \small
    \centering
    \begin{tabular}{@{}r|p{0.79\columnwidth}@{}}
    \toprule

    \multicolumn{2}{c}{QQP (Paraphrase)}   \\  \midrule
    \multirow{2}{*}{Cond} & what is java programming? how to learn java programming language? \\ \hline
    Target                     & how do i learn a computer language like java? \\ \hline
    \multirow{2}{*}{\ModelName{}}  
                             & how can i learn java programming language?  \\
    \cline{2-2}
                             & how should i learn java programming to begin? \\ 
    \midrule[.8pt]
                       \multicolumn{2}{c}{Wiki-Auto (TS)}   \\  \midrule
    \multirow{5}{*}{Cond} & the 7 july 2005 london bombings, often referred to as 7 / 7, were a series of coordinated islamist terrorist suicide attacks in london, england, that targeted commuters travelling on the city's public transport system during the morning rush hour. \\ \hline
    \multirow{4}{*}{Target}  & the 7 july 2005 london bombings ( also called 7 / 7 ) were suicide bomb attacks aimed at london's public transport system during the morning rush hour. \\ \hline
    \multirow{4}{*}{\ModelName{}}  
                             & the 7 july 2005 london bombings were often referred to as 7 / 7. \\
    \cline{2-2}
                             & the 7 july 2005 london bombings, often referred to as 7 / 7, were a series of coordinated started.  \\
  \midrule[.8pt]
                       \multicolumn{2}{c}{Quasar-T (QG)}   \\  \midrule
    \multirow{4}{*}{Cond} & the pound and the euro also took major hits against the yen, indicating investors are losing confidence in their carry trades with the japanese currency. \\ \hline
    {Target}  & what is the japanese currency? \\ \hline
    \multirow{2}{*}{\ModelName{}}  
                             & what is the japanese currency \\
    \cline{2-2}
                             & what is the japanese currency \\
  \midrule[.8pt]
    \multicolumn{2}{c}{CCD (DG)}   \\  \midrule
    {Cond} & great article. thanks for posting \\ \hline
    {Target}  & thanks for reading! \\ \hline
    \multirow{2}{*}{\ModelName{}}  
                             &  happy to help. \\
    \cline{2-2}
                             &  no problem, i was happy. it's awesome. \\
    \bottomrule
    \end{tabular}
    \caption{
      The text generation results for four tasks in the test sets.
      Cond indicates conditional text.
    }
    \vspace{-4mm}
    \label{tab: Generation}
  \end{table}

    \vspace{-1mm}
    \paragraph{Case Study.}

      Four randomly selected samples from each of the four datasets were shown in Table~\ref{tab: Generation}.
      As we can see, the generated results were well controlled by conditional texts.
      \ModelName{} was able to generate different samples under different random seed conditions, except for Quasar-T which included the same target texts for different conditions.
      For CCD, the expression ``\emph{no problem}'' in the second response was not suitable. 
      More efforts should be made for better context understanding.